\documentclass{article}

\PassOptionsToPackage{numbers,sort&compress}{natbib}

\usepackage[preprint]{neurips_2026}

\workshoptitle{Workshop for Autonomous Machine Learning Research}

\usepackage[utf8]{inputenc}
\usepackage{amsfonts}
\usepackage{nicefrac}
\usepackage{microtype}
\usepackage[T1]{fontenc}
\usepackage{microtype}
\usepackage{booktabs}
\usepackage{xcolor}
\usepackage{framed}
\usepackage{xurl}
\usepackage{enumitem}
\usepackage{hyperref}
\usepackage{amsmath}
\usepackage{graphicx}
\usepackage{amssymb}

\title{Autonomous Research Project Management as an Agent Skill: A Case Study in Exact Spectral Spatial Regression}

\author{%
  Alexander Chen$^{a}$, Jeffrey Meng$^{a}$, Bram Hoex$^{a,b}$, Tong Xie$^{a,b}$\thanks{Corresponding author: \texttt{tong.xie@unsw.edu.au}} \\
  $^{a}$School of Photovoltaic and Renewable Energy Engineering\\ 
  University of New South Wales \\
  Kensington, NSW, Australia \\
  $^{b}$GreenDynamics \\
  Kensington, NSW, Australia 
}

\begin{document}

\maketitle

\begin{center}
\small
\textit{This manuscript was originally prepared for the Autonomous
Machine Learning Research (AutoMLR) 2026 workshop and therefore retains
the workshop's prescribed section structure and formatting. This
preprint is released to make the work publicly accessible beyond
the workshop.}
\end{center}

\begin{abstract}
This work presents an end-to-end demonstration of autonomous machine learning research conducted by an agent skill on consumer hardware. The paper's primary qualifying result -- autonomously formulated, implemented, and benchmarked by the agent -- establishes that exact Kernel Ridge Regression (KRR) on regular spatial grids can be solved in closed form in $O(N \log N)$ time and $O(N)$ memory. Using Block Circulant with Circulant Blocks (BCCB) embeddings and 2D-FFT spectral shrinkage, this approach eliminates the prohibitive $O(N^3)$ computational bottleneck of dense kernel methods without relying on basis approximations. Evaluated on 2,005 monthly fields of NOAA Kaplan SST v2 climate anomalies ($36 \times 72$ grid), the spectral solver matches a dense floored-torus reference within $2.62 \times 10^{-12}$ relative infinity-norm precision, achieving masked reconstruction RMSEs of $0.082\text{--}0.088$ ($\sim 14\%$ of field standard deviation). The autonomous workflow systematically characterised the non-periodic free-boundary gap (0.95 for Matérn-3/2, 0.12 for RBF), diagnosed localised split-conformal coverage breakdowns under spatial autocorrelation ($0.739\text{--}0.836$ at halo boundaries against a 0.90 target), and pre-registered a transfer-forecasting hypothesis that was rigorously refuted across 0/4 horizons.  

The research was autonomously executed by DeepSeek V4 Flash, orchestrated by our agent skill suite within DeepSeek Harness (DSH). Experiments were executed on CPU-only hardware (Apple M2 Pro; 78.7 s solver time, 1.57 GB peak RSS). Long-horizon state was decoupled into a file-based epic- and issue-tracking substrate. Across 74 sub-agent sessions, the agent demonstrated closed-loop scientific resilience: routing two failed hypothesis review gates back to literature retrieval, patching bootstrap indexing bugs, and executing with only four discrete human steering events. Finally, we reflect on autonomous research governance, arguing that scientific credibility requires inspectable state, falsifiable review gates, and transparent reporting of negative results, urging the machine learning community to favour agent-accessible structured formats over static PDF manuscripts.
\end{abstract}

\section{Autonomous research result}
\label{page:part1}

\subsection{Introduction}
\label{sec:intro}

Kernel ridge regression (KRR) is a workhorse for spatial prediction in AI for climate \cite{rolnick2022tackling}, but dense solves scale prohibitively as $O(N^3)$ time and $O(N^2)$ memory. On regular grids, circulant embedding \cite{dietrich1997fast,wood1994statistical} and FFT spectral shrinkage keep the solve exact for the periodised torus kernel \cite{gray2006toeplitz}. While 2D circulant embeddings and FFT diagonalisations are established in harmonic analysis, their closed-form application to regular-grid spatial regression with exact PCG boundary preconditioning remains under-characterised. We contribute an exact, CPU-bounded KRR formulation for regular grids, a leakage-verified evaluation protocol, and an empirical benchmark against standard approximations (Nystr\"om, RFF).

\subsection{Related work}
\label{sec:rel}

\textbf{Structured and randomised kernel methods.} On regular grids, stationary kernel Gram matrices are BCCB, enabling exact Gaussian field simulation \cite{dietrich1997fast,wood1994statistical}; we make the regression solve exact via FFT spectral shrinkage \cite{gray2006toeplitz}. Approximate alternatives trade exactness for scale: random Fourier features (RFF) \cite{rahimi2007random}, Nystr\"om \cite{williams2001using,drineas2005nystrom}, Fastfood \cite{le2013fastfood}, KISS-GP \cite{wilson2015kernel}, GPyTorch \cite{gardner2018gpyTorch}, spectral-GP \cite{lazaro2010sparse}, and SPDE/INLA \cite{lindgren2011explicit}. Unlike sketch-circulant KRR on non-grid data \cite{yin2019sketch}, our grid formulation guarantees exactness against dense references ($\sim\!10^{-12}$ relative; Section~\ref{sec:p0}).

\textbf{Uncertainty quantification and forecast evaluation.} Split-conformal intervals require exchangeability; for serially dependent residuals, we calibrate calendar-month blocks from the 60 terminal in-train years \cite{chernozhukov2018exact,gibbs2021adaptive,zaffran2022adaptive}. For autocorrelated errors, blocked cross-validation and block bootstrap resampling provide robust inference \cite{diebold1995comparing,kunsch1989jackknife,bergmeir2012use,roberts2017crossvalidation}; with $n{=}7$ test blocks, we report year-block bootstrap MSE difference CIs. In seasonal forecasting, linear inverse models \cite{penland1993prediction} and NMME ensembles \cite{barnston2012skill,kirtman2014nmme} bound single-model skill under the spring predictability barrier \cite{webster1992monsoon}, while deep models \cite{ham2019deep,zhou2022hybrid} demand GPU resources. Our transfer arm (Section~\ref{sec:t3}) is a leakage-verified benchmark against persistence, climatology, and AR(1), not an operational forecast claim.

\subsection{Method: exact torus spectral KRR}
\label{sec:method}

\textbf{Embedding and training.} Circular row/column distances on an $H\times W$ regular grid yield a BCCB Gram matrix $K=U\,\mathrm{diag}(\hat{c})\,U^*$ diagonalised by the unitary 2D-DFT $U$, where $\hat{c}=\mathcal{F}_{2}(c)$ and $c$ is $K$'s first column \cite{gray2006toeplitz,dietrich1997fast}. Ridge regression $\alpha=(K+\lambda I)^{-1}y$ reduces to $\alpha=\mathcal{F}^{-1}_{2}\!\bigl(\mathcal{F}_{2}(y)/(\widehat{c}+\lambda)\bigr)$ in two FFTs and one pointwise division, scaling as $O(N\log N)$ time and $O(N)$ space; non-torus kernels (e.g. RBF) with small negative BCCB eigenvalues are floored at zero for PSD projection.
\textbf{Boundary gap and masked PCG.} Because BCCB wraps domain edges, the spectral solve is exact for the embedded-torus kernel, leaving an empirical free-boundary gap (0.95 for Mat\'ern-3/2, 0.12 for RBF). For genuinely masked training, we solve the exact free-boundary system via preconditioned conjugate gradients (PCG~\cite{chan1996conjugate}; exact BTTB matvec via the $(2N-1)$-mirror circulant, preconditioned by the floored-torus spectral inverse to $10^{-8}$ tolerance; Section~\ref{sec:t1b}).

\subsection{Data and evaluation protocol}
\label{sec:data}

\textbf{Real data.} NOAA PSL Kaplan SST v2 monthly anomalies \cite{kaplan1998analyses,noaapsl2023kaplansst}: 2,005 fields (1856-01..2023-01, $36\times72$, $5^\circ$ grid); compute is CPU-only. \textbf{Missing data.} 53.4\% of cells carry sentinels (polar caps rows 0--5 and 30--35; land columns); imputed by valid-cell monthly means for training, but strictly excluded from all evaluation targets. \textbf{Valid-only evaluation.} Every metric (RMSE, coverage, skill) evaluates valid cells only; denominators exclude missing cells to prevent dilution. \textbf{Splits.} Train $[0,1920)$ (1856-01..2015-12); test $[1920,2004)$ (2016-01..2022-12; 84 months); month 2004 (2023-01) excluded.
\textbf{Protocol.} (i) masked holdout on test fields: random 10/30/50\% and polar halo ($w{=}1,2,4$) masks; (ii) calendar-month split-conformal intervals (Section~\ref{sec:t1a}), diagnosed per mask on global, seam (2016--2017), and decay (2016) subsets (target 0.90) \cite{chernozhukov2018exact,gibbs2021adaptive,zaffran2022adaptive}; (iii) 5-fold temporal-block CV for functional regression \cite{bergmeir2012use}; (iv) future-blind transfer forecast evaluated via year-block bootstrap (7 blocks, 2,000 resamples) \cite{kunsch1989jackknife}; and (v) coordinate-only pooled KRR baselines (exact, Nystr\"om, RFF) \cite{rahimi2007random,williams2001using,pedregosa2011scikit}.

\subsection{Experiments and results}
\label{sec:results}

\subsubsection{P0: torus exactness on a real field}
\label{sec:p0}
On field 2016-01 ($\lambda{=}10^{-3}$), the spectral solve matches the dense floored-torus reference within $2.62\times10^{-12}$ (Mat\'ern-3/2) and $1.12\times10^{-12}$ (RBF) relative $\ell_\infty$-norm -- passing the $10^{-6}$ gate (free-boundary gap 0.95 / 0.12).

\subsubsection{Reconstruction under leakage-aware masks (T1a)}
\label{sec:t1a}
On the 84 test months ($\lambda{=}10^{-4}$; Table~\ref{tab:t1a}), Mat\'ern-3/2 achieves RMSE 0.0816--0.0880 ($\sim$14\% of anomaly std 0.6001), outperforming RBF (0.2356--0.3155). Random-vs-halo gap is $+0.0040$ (random 30\% 0.0856 vs.\ halo $w{=}2$ 0.0816). Global coverage (target 0.90) reaches 0.877--0.898 (Mat\'ern-3/2) and 0.816--0.893 (RBF). Localised violations are disclosed explicitly: Mat\'ern-3/2 halo $w{=}2$ decay drops to 0.747, halo $w{=}4$ seam/decay to 0.836 / 0.739, and RBF halo $w{=}2$ global to 0.816. Conformity scores are $|y-\hat{y}|$ on valid cells of terminal in-train fields 1860--1919 of matching calendar months; $\lambda$ is tuned on 1910--1919 with the test window held out.

\begin{figure}[t]
\centering
\includegraphics[width=0.98\linewidth]{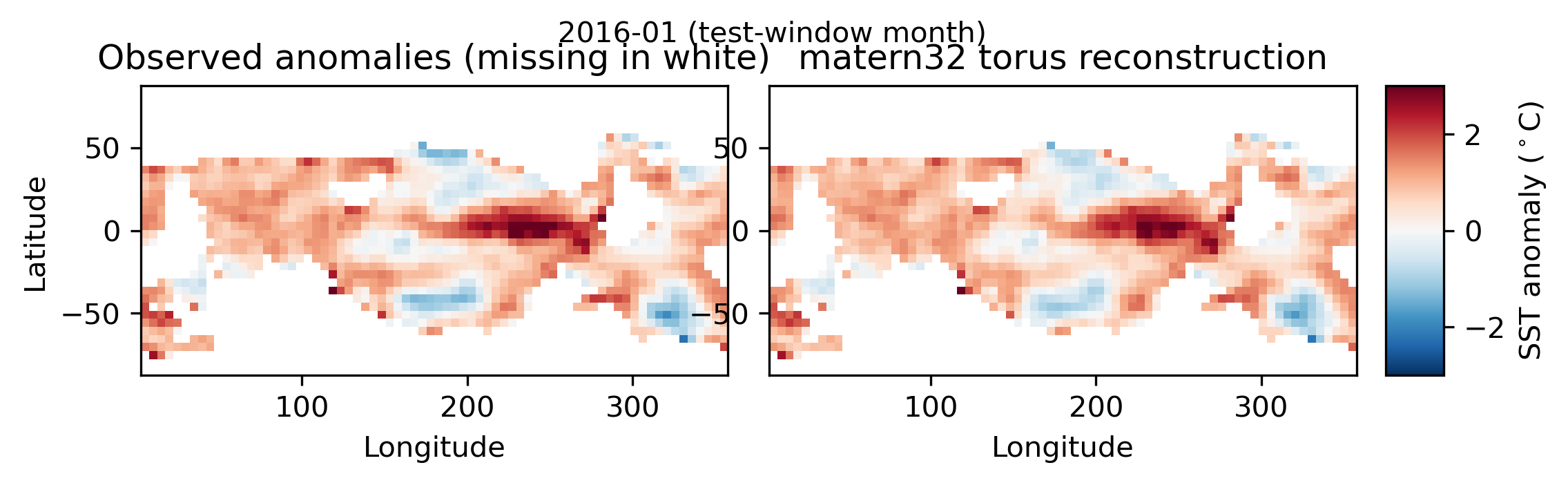}
\vspace{-1.5mm}
\caption{2016-01 test-window field: observed anomalies (missing cells in white) vs.\ Mat\'ern-3/2 torus reconstruction ($\lambda{=}10^{-4}$, Section~\ref{sec:t1a}).}
\label{fig:recon}
\end{figure}

\begin{table}[b]
\centering
\begin{minipage}[t]{0.48\linewidth}
\centering
\caption{T1a masked reconstruction RMSE and valid-cell coverage (target 0.90) on 84 test months.}
\label{tab:t1a}
\vspace{0.4ex}
\footnotesize
\setlength{\tabcolsep}{2.6pt}
\begin{tabular}{lcccc}
\toprule
& \multicolumn{2}{c}{\textbf{Matérn-3/2}} & \multicolumn{2}{c}{\textbf{RBF}} \\
\cmidrule(lr){2-3} \cmidrule(lr){4-5}
\textbf{Mask} & \textbf{RMSE} & \textbf{Cov} & \textbf{RMSE} & \textbf{Cov} \\
\midrule
random 10\%   & 0.0843 & 0.898 & 0.2525 & 0.873 \\
random 30\%   & 0.0856 & 0.892 & 0.2411 & 0.893 \\
random 50\%   & 0.0872 & 0.886 & 0.2356 & 0.892 \\
halo $w{=}1$  & 0.0827 & 0.896 & 0.2789 & 0.856 \\
halo $w{=}2$  & 0.0816 & 0.897 & 0.2839 & 0.816 \\
halo $w{=}4$  & 0.0880 & 0.877 & 0.3155 & 0.824 \\
\bottomrule
\end{tabular}
\end{minipage}%
\hfill
\begin{minipage}[t]{0.49\linewidth}
\centering
\caption{T3 transfer RMSE by lead (84 test months; MSSS skill vs.\ climatology). \textbf{REFUTED} (0/4).}
\label{tab:t3}
\vspace{0.4ex}
\footnotesize
\setlength{\tabcolsep}{2.2pt}
\begin{tabular}{lccccc}
\toprule
$h$ & \textbf{Transfer} & \textbf{Persist} & \textbf{Climat} & \textbf{AR(1)} & \textbf{Skill} \\
\midrule
1  & 0.268 & 0.251 & 0.773 & 0.241 & 0.880 \\
3  & 0.565 & 0.562 & 0.773 & 0.508 & 0.466 \\
6  & 0.796 & 0.907 & 0.773 & 0.766 & $-0.061$ \\
12 & 0.923 & 1.154 & 0.773 & 0.879 & $-0.427$ \\
\bottomrule
\end{tabular}
\end{minipage}
\end{table}

\subsubsection{Genuinely masked training (T1b)}
\label{sec:t1b}
Exact free-boundary PCG (Section~\ref{sec:method}) at 10/30/50\% observed converges to $10^{-8}$ across all runs: mean iterations 245--281 (span 233--291), residuals $4.69\text{--}9.81\times10^{-9}$, masked RMSE 0.529--0.541; full-field spectral T1a remains more accurate.

\subsubsection{Nino3.4 functional index (T2)}
\label{sec:t2}
\begin{figure}[t]
\centering
\begin{minipage}[b]{0.52\linewidth}
  \centering
  \includegraphics[height=2.9cm, keepaspectratio]{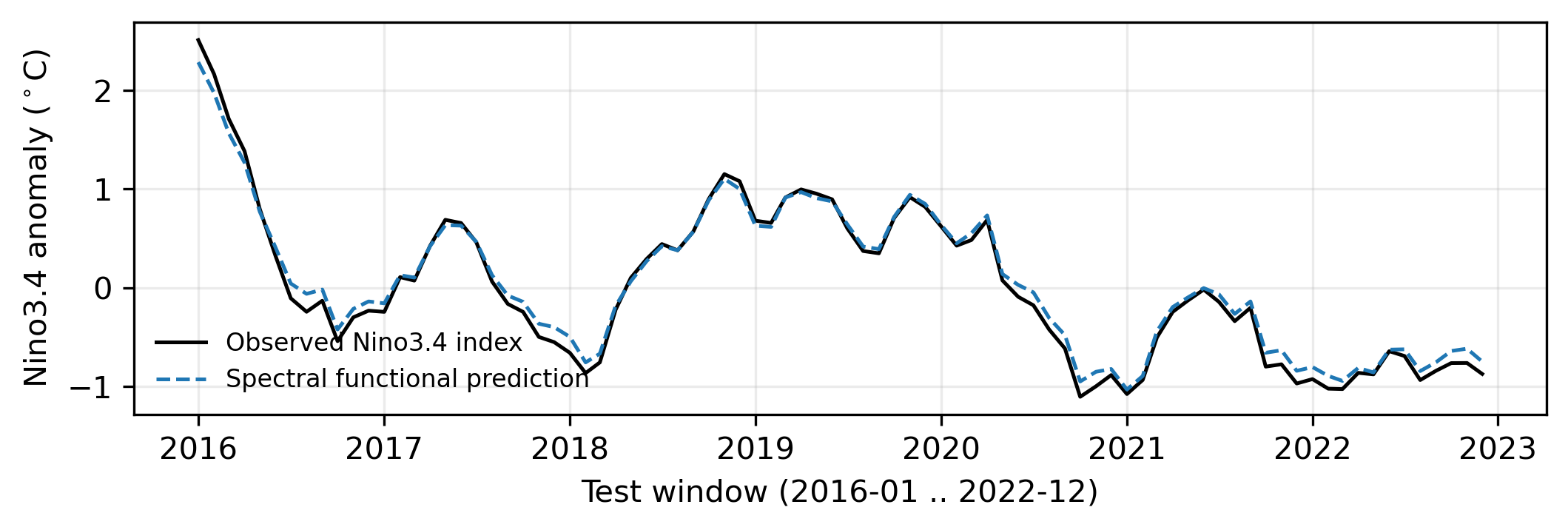}
  \par\vspace{2pt}{\footnotesize (a) Nino3.4 functional tracking}
\end{minipage}\hfill
\begin{minipage}[b]{0.46\linewidth}
  \centering
  \includegraphics[height=2.9cm, keepaspectratio]{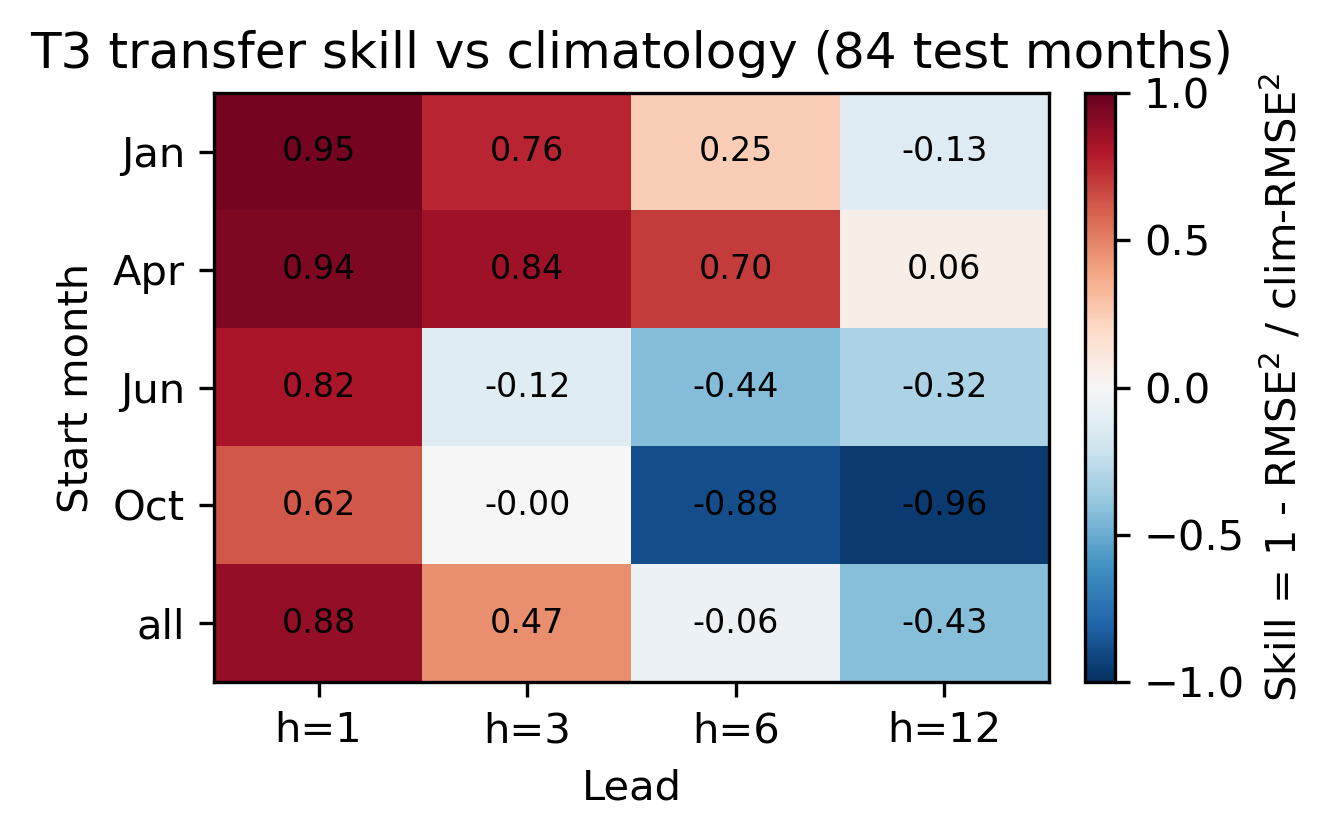}
  \par\vspace{2pt}{\footnotesize (b) Transfer skill vs.\ lead}
\end{minipage}
\vspace{-1.5mm}
\caption{Temporal performance: (a) Nino3.4 index vs.\ spectral box-functional fit ($\lambda{=}10^{-2}$; CV RMSE 0.0856, corr 0.997); (b) T3 transfer skill vs.\ climatology by start month and lead ($n{=}7$ per origin).}
\label{fig:ts}
\end{figure}

Integrating over the Nino3.4 box \cite{trenberth1997definition} via $K(\cdot,\mathbf{1}_{\mathrm{box}})$ achieves RMSE 0.0856 (corr 0.997; Figure~\ref{fig:ts}a) under 5-fold temporal-block CV over all 2,005 months, outperforming direct ridge on raw 2,592-dim features (0.2696, $3.2\times$) and zero baseline (0.7955). This is a method baseline, not an operational forecast claim.

\subsubsection{Transfer forecast (T3): REFUTED under pre-registered win rule}
\label{sec:t3}
Pre-registered win rule: transfer must outperform persistence, climatology, and AR(1) in RMSE and skill with bootstrap-significant margin across horizons. \textbf{Aggregate verdict: REFUTED (0/4 horizons supported; Table~\ref{tab:t3})}. At $h{=}1$, transfer (0.268) loses to persistence (0.251) and AR(1) (0.241); at $h{=}3$, it trails AR(1) (0.508); at $h{=}6$, improvement over persistence (0.796 vs.\ 0.907) is insignificant (95\% CI spans zero, $p{=}0.234$, se 0.197); at $h{=}12$, skill drops to $-0.427$. Bootstrap CI half-widths across comparisons set the minimum detectable MSE margin (observed se 0.0056--0.333); the closest margin at $h{=}1$ is $-0.0137$ (trailing AR(1)), failing significance. Truncation sweeps (2012--2014 cutoffs) maintain $h{=}1$ skill at 0.8795--0.8797. Skill exhibits strong seasonality (Figure~\ref{fig:ts}b): January leads at $h{=}1$ (0.954), April is strongest at $h{\ge}3$ (0.84/0.70/0.06), whereas October degrades rapidly ($0.62 / -0.00 / -0.88 / -0.96$ across $h\in\{1,3,6,12\}$).

\subsubsection{Solve-cost control on a fixed pooled-cell task (B)}
\label{sec:b}
On 2,000 pooled test cells, spectral reconstruction achieves RMSE 0.0999; coordinate-only exact/Nystr\"om/RFF KRR reach 0.820--0.824 while requiring $4.8\times 10^2$ to $1.6\times 10^6$ relative FLOPs. Coordinate baselines sit at field-std level (zero 0.801, mean 0.605, climatology 0.813).

\subsubsection{Resolution sensitivity (R) and scaling (S)}
Coarsening $2\times2$ to $10^\circ$ degrades reconstruction RMSE from 0.0853 to 0.1667 and functional RMSE from 0.0856 to 0.3825 (corr 0.997 to 0.950). Fits remain sub-millisecond through $N{=}10,368$ (0.2--0.5\,ms), requiring five working arrays ($\sim$0.04\,GB for $10^6$ cells). Peak CPU-only resource footprint on Apple M2 Pro: 78.7\,s wall-clock, 1.57\,GB RSS.

\subsection{Limitations and discussion}
\label{sec:limits}
(i) \textbf{Torus exactness scope}: spectral efficiency applies to embedded-torus kernels; free-boundary masked training requires iterative PCG at $\sim$6$\times$ higher error. (ii) \textbf{Transfer claim refuted}: 1-month transfer skill reflects autocorrelation already captured by AR(1); negative skill floor at $h{=}12$. (iii) \textbf{Non-uniform coverage}: localised seam/decay coverage degrades to 0.739--0.837 under autocorrelation. (iv) \textbf{Resolution sensitivity}: spatial coarsening degrades reconstruction. (v) \textbf{Kaplan smoothing}: underlying EOF reconstruction relies on monthly mean imputation for 53.4\% missing entries \cite{kaplan1998analyses}.

\subsection{Conclusion}
Exact circulant-block KRR provides a practical, closed-form CPU-only route on regular grids: torus-exact spectral solves at $O(N\log N)$ time and $O(N)$ memory (78.7\,s / 1.57\,GB) with masked reconstruction errors at $\sim$14\% of field standard deviation. By fully disclosing empirical limitations, conformal breakdowns, and refuted transfer hypotheses, this study provides a transparent, resource-bounded algorithmic benchmark for structured spatial regression.
\clearpage
\section{System Design and Meta-Analysis}
\label{page:part2}

This section provides the system design and meta-analysis of the autonomous research pipeline that produced the results in Section~\ref{page:part1}, detailing the agent architecture, file-based issue-tracking harness, execution loops, human interventions, and verification protocols.

\subsection{Autonomous Research as an Agent Skill}
\label{sec:agent_skill}

Our autonomous research system encapsulates end-to-end scientific workflows within modular, composable \emph{agent skills}. To manage long-horizon execution without conversational drift or context loss, the system employs an Agile-inspired tracking architecture. Research state is decoupled entirely from LLM conversational memory and maintained exclusively via plain Markdown artifacts (\texttt{EPIC.md}, \texttt{ISSUE.md}, and \texttt{comments/*.md}) on the local file system. Skills define deterministic bookkeeping operations over this substrate, ensuring that human steering and multi-turn feedback translate directly into version-controlled issue records.

\subsubsection{Core Skills: Epic and Issue Tracking}

The substrate provides a minimal set of core tracking skills: \texttt{add-epic}, \texttt{add-issue}, \texttt{add-comment}, \texttt{issue-manager}, and \texttt{issue-subagent-orchestration}. 

\textbf{Role Isolation and State Locality.} To prevent cascading hallucinations and state pollution, we enforce strict role isolation. Only the designated \textbf{issue manager} possesses permission to mutate issue metadata (\texttt{ISSUE.md}), record comments, or spawn child processes. Worker sub-agents (such as code generators and reviewers) operate in complete isolation from tracking semantics; they receive self-contained briefs and return structured JSON outputs without visibility into the broader tracking topology. All execution state persists in deterministic file structures: global milestone scope and canonical issue indices reside in \path{EPICS_ROOT/<epic>/EPIC.md}, local task descriptions and acceptance criteria in \path{issues/<issue>/ISSUE.md}, and append-only history and discussion in \path{issues/<issue>/comments/*.md}.

Three properties make this architecture attractive: project trajectory is inspectable and diffable via standard version control; execution state survives resets and agent context truncations; and system claims are auditable against disk artifacts.

\textbf{Reflexive Substrate (System-as-Subject).} Because skills, verification suites, and orchestrator configs are plain-text code artifacts within the workspace, the tracking schema is fully reflexive. An epic's target deliverable can be a scientific paper, an empirical benchmark, or the tracking system's own skill implementation. This uniform representation eliminates the need for separate meta-orchestration frameworks when developing, profiling, or refactoring agent skills.

\textbf{Harness Primitives and Validation Batteries.} The system interfaces with the underlying agent harness (DSH) through six execution primitives: \texttt{subagent} and \texttt{subagent\_fork} for nested process dispatch, \texttt{send\_message} for interactive follow-up, \texttt{interrupt\_agent} for early termination, \texttt{job\_output} / \texttt{job\_list} for output polling, and \texttt{workflow} for scripted concurrency. To ensure substrate integrity, each tracking skill embeds a standalone, Python standard-library validation battery (\texttt{test\_add\_epic.py}, \texttt{test\_add\_issue.py}, \texttt{test\_add\_comment.py}, and \texttt{test\_lifecycle.py}). These suites enforce structural invariants: discovery roots, git cleanliness, file layout conventions, and ASCII encoding.

\subsubsection{Epic Manager Skill: Research Project Manager}

While the core tracking skills define an issue manager responsible for orchestrating worker sub-agents within the scope of a single task, we intentionally omit a generic epic manager. Instead, the epic manager is implemented as a specialised skill tailored to the lifecycle of an end-to-end scientific research project. It coordinates long-horizon research across five sequential phases: Project Scoping (Phase~A), Control Validation and Cost Tracking (Phase~B), Stage-Driven Execution (Phase~C), Staging and Collection (Phase~D), and Synthesis (Phase~E).

In \textbf{Phase~A (Project Scoping)} and \textbf{Phase~B (Control Validation)}, the system establishes invariant state without overcommitting to upfront execution plans. Guided by a user research brief and declarative configuration, the epic manager evaluates plan viability, initialises the project workspace, and creates a designated control issue (\texttt{orchestration-log}) alongside a manifest skeleton (\texttt{project.json}). Crucially, the system eschews bulk, waterfall issue pre-generation. Phase~B validates on-disk issue tracking invariants and initialises a mandatory cost-harvesting ledger to account for token consumption and sub-agent invocations across all downstream stages.

\textbf{Phase~C (Stage-Driven Dispatch)} governs the core research loop over a canonical scientific spine: literature review, hypothesis ideation, experiment planning, execution, analysis, and writeup. To preserve flexibility as scientific findings emerge, issues are instantiated \emph{just-in-time} at the threshold of each stage. The manager dispatches isolated issue-manager sub-agents up to configurable concurrency limits, dual-writing directives to comment threads and managing execution states. Critical transition points -- such as hypothesis selection and post-experiment analysis -- are guarded by formal review gates. External reviewer agents provide criterion-level scores rather than opaque pass/fail decisions; the epic manager evaluates these scores against threshold policies, programmatically deriving verdicts. On failure, the manager orchestrates backward routing, superseding invalid downstream issues, wiping invalidated intermediate artifacts, and re-entering prior stages with structured critique.

Following execution, \textbf{Phase~D (Staging and Collection)} settles the project manifest, harvesting findings, logs, and token costs from terminal research issues. The manager evaluates strict status precedence: loop-exhaustion limits override nominal task completion, explicitly marking stalled research as blocked rather than falsely converged. Finally, \textbf{Phase~E (Synthesis)} compiles the research paper under strict reporting-integrity obligations. An artifact evidence scanner compiles on-disk JSON summaries into a factual evidence preface. The deliverable research manuscript is then refined through an iterative writeup clarity review loop.

\subsubsection{Cross-Harness Portability} 

Although developed and benchmarked on DSH, the system maintains no hard programmatic coupling to DSH internals. Any harness supporting nested sub-agent dispatch and local file system access can run this system.

Porting the system requires only two configuration steps: (1) symlinking the \path{skills/} directory into the target harness's discovery root alongside \path{AGENTS.md}, and (2) adapting the dispatch wrappers in \texttt{issue-subagent-orchestration} and \texttt{research-project-epic-manager} to map onto the target harness's native sub-agent spawning, messaging, and process collection primitives.

\subsubsection{Comparison to Other Systems}

Existing autonomous research platforms broadly divide into end-to-end pipelines~\citep{aiscientistv1,aiscientistv2,deepscientist2025,kosmos2025} and stage-bounded specialists targeting ideation, literature synthesis, or protocol design~\citep{schmidgall2025agentlab,coscientist2026,storm2024,funsearch2024,researchagent2025}. Across both paradigms, state persistence typically relies on transient conversational context, unstructured run directories, or opaque graph checkpoints~\citep{autogen2024,metagpt2024,langgraph2024}. In contrast, our architecture decouples execution state into a deterministic, file-based tracking substrate (\texttt{EPIC.md}, \texttt{ISSUE.md}). Storing project history as plain Markdown renders the scientific trajectory human-inspectable, git-diffable, and editable out-of-band with standard developer tools without database deserialisation.

A second divergence lies in the locus of self-improvement. Prior systems improve at the \emph{object level} via search and re-prompting~\citep{aiscientistv1,funsearch2024}, the \emph{memory level} via cross-run findings stores~\citep{deepscientist2025}, or the \emph{weight level} via policy retraining~\citep{cycleresearcher2025}. Our substrate introduces a representation for \emph{protocol-level} self-improvement: because skills, review rubrics, and orchestration scripts are standard code artifacts, the tracking system can treat the project manager skill itself as the subject of an epic (``system-as-subject''). While our core tracking skills were used to dogfood and develop the project manager skill in this manner, we emphasise that closed-loop, fully autonomous protocol refinement -- where an agent mines prior execution traces to initiate and merge modifications to its own skills without human oversight -- remains an open research problem that we do not claim to solve here.

\subsection{Run-Specific Settings}

This subsection details the experimental configuration, harness execution, tool access, human interventions, and verification protocols that produced the real-data spatial regression study in Section~\ref{page:part1}, instantiated under the architecture established in Section~\ref{sec:agent_skill}.

\subsubsection{Agent and Harness}

The primary orchestrator was DeepSeek V4 Flash, deployed within the DeepSeek Harness (DSH). The agent operated via the modular skill framework detailed in Section~\ref{sec:agent_skill}, assuming the role-isolated \textbf{issue manager} to govern task progression, state mutation, and sub-agent dispatch over local file artifacts without conversational context drift.

\subsubsection{Tools}

The system operated within a local Docker container on an Apple M2 Pro, constrained to 12 CPU cores and $\sim$7~GB of host RAM ($\sim$2~GB container cgroup limit) without GPU acceleration. The runtime environment consisted of Python 3.10 with \texttt{numpy} 1.24.2, \texttt{scipy} 1.10.1, \texttt{scikit-learn} 1.2.1, and \texttt{h5py} (added exclusively for read-only ingestion of gridded NetCDF-4/HDF5 climate observations~\citep{kaplan1998analyses}). PDF rendering utilised \texttt{tectonic} 0.17.0. External web retrieval was mediated via the Tavily MCP \citep{tavily2024}; all scientific computing, feature transformations, and matrix solves executed offline. Over the run, 94 in-window Tavily API calls were executed (12 by the manager, 82 by dispatched sub-agents) primarily during literature grounding.

\subsubsection{Research Loop}

The research loop spanned two consecutive iterations over a total duration of 13~h 52~min. Iteration~1 executed a complete synthetic benchmark pipeline. Following an external steering directive, Iteration~2 (6~h 6~min) re-entered the hypothesis ideation stage to execute a real-data study on NOAA Kaplan SST v2 anomalies~\citep{kaplan1998analyses}:
\begin{itemize}
    \item \textbf{Ideation and Hypothesis Gating:} Proposal ideation executed three iterative critique rounds, halting at the round cap (3/3) due to strict novelty criteria. The subsequent formal hypothesis gate failed twice (exhausting its 2-loop budget). Applying the automated routing protocol, the manager diverted execution to \texttt{literature\_review} for corpus strengthening, incorporating 15 web-verified citations and expanding the concept graph to 492 entities before re-entering ideation and securing gate approval.
    \item \textbf{Execution and Analysis:} Experiment execution evaluated exact spectral KRR, masked PCG reconstruction, functional Nino3.4 regression, and transfer forecasting in 78.7~s of CPU wall-clock time and 1.57~GB peak RSS. During execution checks, sub-agent reviews caught two critical bugs: a calendar-year mapping collapse that rendered bootstrap confidence intervals vacuous, and an array index bias in baseline evaluation; both were corrected in code before gating. The analysis gate subsequently passed on round 1 (median 4/4/4).
    \item \textbf{Synthesis and Writeup:} Paper generation drafted the 4-page manuscript. The writeup clarity gate failed round 1 (scores 3/3), identifying 12 concrete defects (e.g. table overflow, missing acronym expansions, and origin-forecast reporting contradictions), and passed round 2 (scores 4/4) after revisions.
\end{itemize}
Across 74 sub-agent sessions, the run consumed 1,625 LLM invocations totalling 203.3M metered tokens, with prompt caching absorbing 96.7\% of the input volume.

\subsubsection{Human Interventions}

Human intervention during the research loop consisted of four steering events:
\begin{enumerate}
  \item \textbf{Research Brief:} Initialised Phase~A scope for a 4-page PDF targeting an ML algorithmic contribution in AI-for-climate under consumer CPU bounds (12 cores, 7\,GB RAM, $<$90\,min).
  \item \textbf{Iteration-2 Directive:} Rerouted the pipeline back to ideation for higher empirical significance, explicitly permitting dependency additions (\texttt{h5py}) to ingest real observational data after Iteration~1 strictly avoided external packages.
  \item \textbf{Page-Budget Clarification:} Notified the writeup worker that references were exempt from the 4-page limit, preventing an attempt to compress citations into the main text budget.
  \item \textbf{Draft Review Feedback:} Prompted the agent following visual inspection to incorporate missing related works and presentation figures.
\end{enumerate}

\textbf{Post-Generation Formatting and Citation Audits:} Following generation, human-prompted Gemini 3.8 Flash adapted the draft to align with AutoMLR workshop formatting and cross-checked citations. 

\subsubsection{Verification}

Artifact verifiability emerged as an intrinsic property of the issue-tracking harness rather than manual inspection of derivations. Post-generation manual proofreading focused on presentation fidelity, while algorithmic correctness was audited via automated verification suites. To double-check hallucinated content, the human steered an independent post-synthesis re-verification issue (\texttt{verification-method-code-divergence-and-empirical-claims}), which checked:
\begin{itemize}
  \item \textbf{Method-Code Divergence:} Algorithmic writeups (BCCB embedding, mirror PCG) were mapped to implementing functions in \texttt{scripts/}, ensuring the text strictly reflects executed code without hallucinated methodology.
  \item \textbf{Superseded-Number Leakage:} The manuscript was audited against \texttt{EXECUTION\_NOTES.md} to ensure no superseded metrics (e.g., pre-fix T1b iterations or iteration-1 synthetic figures) were inadvertently cited.
  \item \textbf{Claim Traceability:} An assertion suite (\texttt{check\_report.py}) executed value-by-value re-verification, reproducing 77/77 quantitative claims, bootstrap CIs, and compute footprints against \texttt{results/iter2/*.json}.
\end{itemize}

\subsubsection{Significance of Result}

The primary result establishes that exact Kernel Ridge Regression can be solved on regular spatial grids in $O(N \log N)$ time and $O(N)$ memory using Block Circulant with Circulant Block (BCCB) embeddings and 2D-FFT spectral shrinkage~\citep{gray2006toeplitz,dietrich1997fast}. This circumvents the prohibitive $O(N^3)$ computational and $O(N^2)$ memory bottlenecks of dense kernel methods without resorting to approximate basis truncations (such as Nystr{\"o}m or Random Fourier Features)~\citep{williams2001using,rahimi2007random}. Furthermore, the study characterises the exact free-boundary approximation gap (0.95 relative error for Mat{\'e}rn-$3/2$) and demonstrates that commodity CPU hardware can perform exact grid-based inference at scale without GPU infrastructure.

\subsubsection{Interpretability}

Every stage transition, code diff, reviewer score sheet, and execution log was committed sequentially to disk across \texttt{EPIC.md}, \texttt{ISSUE.md}, and append-only \texttt{comments/*.md} threads as described in Section~\ref{sec:agent_skill}. The full operational narrative -- including the two failed hypothesis gates and reviewer bug reports -- remains inspectable and reproducible without inspecting transient model activations.

\subsection{Limitations}

The system design and autonomous execution exhibit four primary limitations:
\begin{itemize}[leftmargin=*]
    \item \textbf{Downstream Empirical Scope:} As detailed in Section 1.6, findings remain bounded by statistical limits (the free-boundary PCG gap, localized conformal under-coverage, and refuted multi-horizon forecasting).
    \item \textbf{Algorithmic Scaling and Negative Falsification:} In an external exploratory run tasked with resolving the boundary gap via DST-II Woodbury solves and randomized-Nystr\"om preconditioning, the agent's pre-registered review gates halted execution when the Nystr\"om preconditioner stalled ($0.135\times$ speedup vs.\ PCG) and full-support boundary effects exceeded the $\sim$2\,GB container cap. While confirming review-gate falsification without hallucinated convergence, this highlights agent vulnerabilities to complex linear-algebra trade-offs under tight container budgets.
    \item \textbf{Residual Human Steering:} Execution was not zero-shot, requiring four discrete human steering events (scoping, authorising \texttt{h5py}, budget clarification, and presentation critique) alongside post-generation editorial auditing.
    \item \textbf{Unrealised Autonomous Protocol Refinement:} Although the file-based tracking substrate supports reflexive tracking of its own skills as deliverables, closed-loop self-refinement without human steering remains an open challenge.
\end{itemize}
\clearpage
\section{Broader Impact}
\label{page:part3}

\paragraph{The future of autonomous research.} 

Prior work has demonstrated that autonomous agents can produce workshop-calibre papers for under \$15 per manuscript~\citep{aiscientistv1,aiscientistv2,deepscientist2025}. Our work builds on this trajectory, establishing that end-to-end scientific workflows can be orchestrated dynamically on modular, agent-skill substrates.

Such decreases in the cost of autonomous research introduce a stark divergence between digital hypothesis velocity and physical experimental throughput. While autonomous agents can formulate hypotheses, optimise execution code, and synthesise statistical analyses at near-zero marginal cost, physical validation -- spanning chemical synthesis, materials characterisation, and in vitro biological assays -- remains strictly bottlenecked by reaction kinetics, mass transport, and physical laboratory automation~\citep{szymanski2023}.

\paragraph{The generation-verification asymmetry.} 

The proliferation of cheap compute and frontier reasoning APIs has democratised access to automated discovery~\citep{aiscientistv1,schmidgall2025agentlab}, driving unprecedented submission volumes across major scientific venues. Concurrently, analyses of peer reviews at {ICLR} 2025 and {Nature Communications} found that 12\% to 20\% of reviews were generated by {LLMs}~\citep{shen2026detecting}.

Crucially, recent empirical audits of autonomous discovery frameworks~\citep{hiddenpitfalls2025,llmreval2025} highlight that ungrounded LLM-as-a-judge pipelines cannot yet serve as trustworthy proxies for conventional scientific peer review. They uncovered severe verifiability failures across all baseline agents; citation hallucination, method-code divergence and unreproducible empirical claims. Developing robust, automated verification protocols remains an open challenge that grows increasingly urgent as synthetic paper output accelerates.
 
\paragraph{Leaving the manuscript PDF behind.} 

Although end-to-end automated verification remains an open problem, we argue that sunsetting the traditional PDF manuscript is a first step in the right direction. Standard page limits incentivise authors to construct artificially clean narratives, systematically suppressing negative results, null hypotheses, and anomalous edge cases~\citep{rosenthal1979file,fanelli2012negative}. Furthermore, in an ecosystem where automated crawlers and synthesis agents ingest the vast majority of scientific literature, publishing into a static PDF is a redundant step that introduces extraction errors and severs execution provenance~\citep{lo2020s2orc,blecher2023nougat}.

Instead, the research community should align its publication venues and incentive structures around three key reforms:
\begin{enumerate}
    \item \textbf{Containerised Execution Deliverables.} The primary scientific artifact should be machine-readable execution objects -- such as structured JSON-LD schemas, hermetic runtime environments, and reproducible execution protocols deposited in decentralised archives~\citep{soiland2022packaging,agentrxiv2025}. These formats allow audit agents to independently extract hyperparameters, re-execute pipelines, and verify intermediate reasoning traces without lossy text parsing, thereby facilitating deterministic verification protocols such as claim-to-evidence trace graphs~\citep{ledger2026}.
    \item \textbf{Dataset-First Evaluation Rubrics.} Scientific venues should transition peer review to focus on dataset quality and empirical significance, realigning academic prestige with the true empirical bottleneck of scientific discovery~\citep{sambasivan2021everyone}. Review criteria should prioritise: (i)~data provenance, noise floor characterisation, and physical hardware calibration rigour; and (ii)~empirical information gain ($\Delta$), quantifying the degree to which a newly collected dataset updates prior predictive distributions over competitive baselines.
    \item \textbf{Granular Micro-Citations.} Citation attribution should move beyond coarse, paper-level citations toward granular, artifact-level micro-citations~\citep{martone2014data,wilkinson2016fair}. As autonomous agents ingest, transform, and synthesise downstream findings, algorithmic credit assignment must propagate back to the primary human researchers who physically collected raw specimens, calibrated hardware, and curated the ground-truth empirical datasets underpinning the models.
\end{enumerate}

\clearpage
\bibliographystyle{unsrtnat}
\bibliography{references}

\clearpage
\appendix

\section{Reproducibility Statement}

The primary result evaluates publicly available NOAA Kaplan SST v2 anomalies using offline Python tooling (\texttt{numpy}, \texttt{scipy}, \texttt{scikit-learn}, and \texttt{h5py}). All experiments are strictly CPU-bound, requiring 78.7\,s wall-clock time and 1.57\,GB peak RSS within a local Docker container on an Apple M2 Pro (12 cores, $\sim$2\,GB memory limit). Reproduction and auditability follow the verification steps in Section~\ref{page:part2}, backed by \texttt{validate\_execution.py}, future-blind temporal splits (1856--2015 train, 2016--2022 test), dense Cholesky numerical equivalence within $2.62 \times 10^{-12}$, and metric traceability to programmatic keys in \texttt{results/iter2/*.json}.

Our codebase -- Research Epic Manager Agent Skills suite can be found at \url{https://github.com/alexchen5/research-epic-manager} -- which also includes the run artifacts of Section~\ref{page:part1}. 

\section{Disclosure Statement}

\paragraph{Agent(s) / model(s) used}
The pipeline was orchestrated by DeepSeek V4 Flash within the DeepSeek Harness (DSH) using our research system skills suite across 74 sub-agent sessions, with external literature search mediated by Tavily MCP. Gemini 3.8 Flash was used post-generation for editorial changes, including citation auditing.

\end{document}